\documentclass[letterpaper, 10 pt, conference]{ieeeconf}  

\IEEEoverridecommandlockouts                              

\usepackage[utf8]{inputenc}
\usepackage[OT1]{fontenc}

\makeatletter
\let\NAT@parse\undefined
\makeatother
\usepackage[numbers]{natbib}

\usepackage{hyperref}
\usepackage{amsmath} 
\usepackage{amssymb}  
\usepackage{mathtools}
\usepackage{cleveref}
\usepackage{bm}
\usepackage{physics}
\usepackage{comment}
\usepackage{xcolor}
\usepackage{siunitx}
\usepackage{etoolbox}
\makeatletter
\patchcmd{\@makecaption}
  {\scshape}
  {}
  {}
  {}
\makeatletter
\patchcmd{\@makecaption}
  {\\}
  {.\ }
  {}
  {}
\makeatother

\DeclareMathOperator*{\argmin}{arg\,min}

\title{\LARGE \bf
Nonlinear Model Predictive Control for Quadrupedal Locomotion \\
Using Second-Order Sensitivity Analysis
}

\author{Dongho Kang, Flavio De Vincenti, Stelian Coros
\thanks{This work has received funding from the European Research Council (ERC) under the European Union’s Horizon 2020 research and innovation programme (grant agreement No. 866480).}
\thanks{The authors are with the Computational Robotics Lab in ETH Zurich, Switzerland. 
{\tt\small \{kangd, dflavio, scoros\}@ethz.ch}}
\thanks{The first two authors contributed equally to this work.}
}

\begin{document}

\maketitle
\thispagestyle{empty}
\pagestyle{empty}

\newcommand{\DK}[1]{{\bf\textcolor{red}{DK: #1}}}
\newcommand{\FDV}[1]{{\bf\textcolor{ForestGreen}{FDV: #1}}}
\newcommand{\SC}[1]{{\bf\textcolor{blue}{SC: #1}}}

\newcommand{\R}{\mathbb{R}}

\newcommand{\X}{\mathbf{X}}
\newcommand{\U}{\mathbf{U}}
\newcommand{\Se}{\mathbf{S}}
\newcommand{\q}{\mathbf{q}}
\newcommand{\refe}{\mathrm{ref}}

\newcommand{\x}{\mathbf{x}}
\newcommand{\p}{\mathbf{p}}
\renewcommand{\u}{\mathbf{u}}

\newcommand{\G}{\mathbf{G}}
\renewcommand{\J}{\mathcal{J}}

\renewcommand{\r}{\mathbf{r}}
\renewcommand{\s}{\mathbf{s}}
\newcommand{\eeq}{\overset{!}{=}}

\begin{abstract}

We present a versatile nonlinear model predictive control (NMPC) formulation for quadrupedal locomotion. 
Our formulation jointly optimizes a base trajectory and a set of footholds over a finite time horizon based on simplified dynamics models.
We leverage second-order sensitivity analysis and a sparse Gauss-Newton (SGN) method to solve the resulting optimal control problems.
We further describe our ongoing effort to verify our approach through simulation and hardware experiments.
Finally, we extend our locomotion framework to deal with challenging tasks that comprise gap crossing, movement on stepping stones, and multi-robot control.
\end{abstract}

\section{Introduction}

Model predictive control (MPC) is a powerful tool for enabling agile and robust locomotion skills on legged systems.
Its capability of handling flying phases while rejecting disturbances enhances the maneuverability \cite{Kim2019HighlyDQ, Ding2019RealtimeMP} and mobility of quadruped robots \cite{kalakrishnan2010fast}. 

Standard MPC implementations are concerned with solving finite-horizon optimal control problems (OCPs) at a real-time rate. 
This process comes with a high computational cost that defies online execution.
Current existing MPC methods for quadrupedal locomotion tackle this challenge through careful software designs and high-performance, parallel implementations \cite{Farshidian2017RealtimeMP, Neunert2018WholeBodyNM, Bledt2019ImplementingRP}. In addition, they adopt simplified dynamics models to reduce computational burdens: one common simplification is pre-defining the footholds with heuristics-based methods \cite{Kim2019HighlyDQ, Ding2019RealtimeMP, Carlo2018DynamicLI} that can restrict the range of achievable motion and the capability to reject external disturbances. 

In this paper, we present a versatile nonlinear MPC (NMPC) strategy that jointly optimizes a base trajectory and a sequence of stepping locations. We describe the system dynamics as a function of a control input vector evolving over a time horizon and a time-invariant set of footholds. We solve the resulting OCP using a second-order numerical solver \cite{Zehnder2021SGNSG} that leverages sensitivity analysis (SA) \cite{Zimmermann2019OptimalCV, devincenti2021control, bern2017fabrication, bern2019trajectory} to compute the exact values of the required derivatives efficiently. This approach significantly improves the robustness of the controller while ensuring real-time execution. Moreover, our formulation is easily adaptable to various nonlinear models and quadrupedal locomotion scenarios.

In the following sections, we provide the mathematical formulation of our method. Furthermore, we describe two examples based on different nonlinear dynamics models compatible with our framework. Finally, we present our preliminary results verifying our approach and applying it to various locomotion control tasks.

\begin{figure} 
    \centering
    \includegraphics[width=\linewidth]{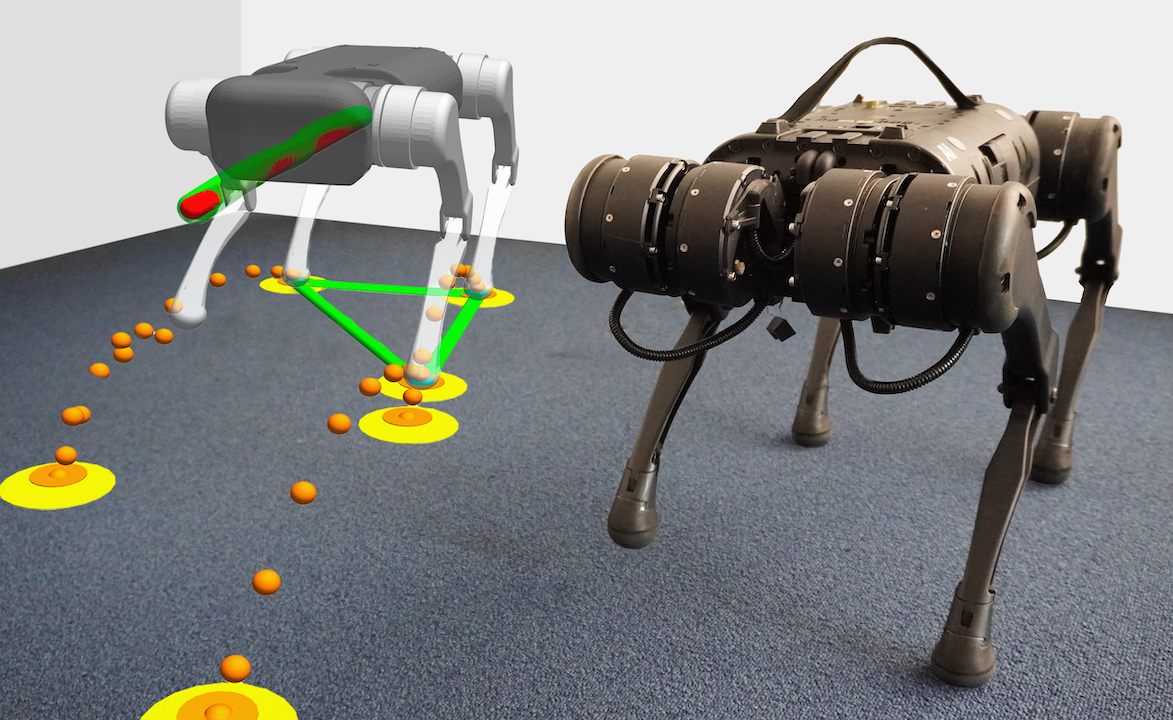}
    \caption{
    Our MPC-based locomotion controller in action on a simulated \emph{Unitree A1} robot \textbf{(left)} and a real one \textbf{(right)}. Given kinematically generated references (\textbf{red curve} and \textbf{yellow circles}), our planner generates optimal base trajectory (\textbf{green curve}) and footholds (\textbf{orange circles}) that are dynamically feasible.}
    \label{fig:teaser}
\end{figure}
\section{Nonlinear MPC}

In this section, we describe our optimal control framework based on second-order SA.
Subsequently, we formulate a model-agnostic OCP for quadrupedal locomotion, where the optimization variables include system states and stepping locations.
Finally, we provide two examples applying our formulation to nonlinear dynamics models; namely, the variable-height inverted pendulum and the single rigid body model.

\subsection{Framework}
\label{sec:framework}

We express the discrete-time dynamics of a system through an implicit function
\begin{equation}\label{eqn:dynamics-implicit}
    \G_k(\x_k,\, \x_{k+1}, \u_k,\, \p) = \mathbf{0}_n \,,
\end{equation}
where $\x_k\in\R^n$ and $\u_k\in\R^{m}$ denote the system state and control input vectors at time step $k$, $\G_k$ is a differentiable function capturing the system evolution at time step $k$, and $\mathbf{0}_n\in\mathbb{R}^n$ is the $n$-dimensional zero vector. In this formulation, the dynamics further depend on a time-invariant vector of parameters $\p\in\R^p$: while $\u_k$ affects the system only at time step $k$, $\p$ does so for multiple time steps. In our application to locomotion control, $\p$ represents a set of footholds to be stepped on over a time horizon (see \Cref{sec:locomotion-control}).

We define the stacked state vector $\X \in \mathbb{R}^{Nn}$ and the stacked input vector $\U \in \mathbb{R}^{Nm+p}$ as follows:
\begin{align*}
    \X &\coloneqq \begin{bmatrix}
    \x_1^\top & \x_2^\top & \dots & \x_N^\top
    \end{bmatrix}^\top \,, \\
    \U &\coloneqq \begin{bmatrix}
    \u_0^\top & \u_1^\top & \dots & \u_{N-1}^\top & \p^\top
    \end{bmatrix}^\top \,, 
\end{align*}
where $N$ denotes the time horizon. Additionally, given a measurement $\x_0$ of the current state of the system, we define the stacked dynamics constraint function as
\begin{equation}
\label{eq:residuals}
\G(\X, \U) \coloneqq \begin{bmatrix}
    \G_0^\top & \G_1^\top & \dots & \G_{N-1}^\top
    \end{bmatrix}^\top\,.
\end{equation}
Then, we can define a finite-horizon OCP for the system \eqref{eqn:dynamics-implicit} 
\begin{align}
\begin{split}
\label{eq:constrained-optimal-control}
    \min_{\X,\,\U} \quad &\J(\X,\,\U) \, \\
    \text{s.t.} \quad &\G(\X, \U) = \mathbf{0}_{Nn} \,,
\end{split}    
\end{align}
where $\mathcal{J}(\X,\,\U)$ is a cost function that depends on the stacked state and input vectors. 

If an explicit function $\mathbf{g}_k$ such that $\x_{k+1} = \mathbf{g}_k(\x_k,\,\u_k,\,\p)$ is available, 
we can define $\G_k \coloneqq \x_{k+1} - \mathbf{g}_k(\x_k,\,\u_k,\,\p)\,$ to adapt the system dynamics to the form \eqref{eqn:dynamics-implicit}.
However, we note that \eqref{eqn:dynamics-implicit} is general enough for cases where an explicit form of the dynamics equation does not exist. For instance, if the dynamics of a system are defined as:
\begin{equation}
    \x_{k+1} \coloneqq \x^\ast = \argmin_{\x} E_k(\x, \x_{k}, \u_{k})\,,
\end{equation}
where $E_k$ is the energy function of the system at time step $k$, then they cannot be made explicit \cite{bern2017fabrication, bern2019trajectory}. Nevertheless, we can define an implicit function $\G_k \coloneqq \frac{\partial E_k}{ \partial \x}$ that is equal to zero for $\x^\ast$ that minimizes $E_k$.

Under mild assumptions, \eqref{eqn:dynamics-implicit} implies that there is a map between $\X$ and $\U$, i.e., $\X(\U)$, although the map may not have an analytic form. Therefore, we can convert \eqref{eq:constrained-optimal-control} into the following unconstrained minimization problem
\begin{equation}
\label{eq:unconstrained-optimal-control}
    \min_{\U} \quad \J(\X(\U),\,\U) \,.
\end{equation}

We find the optimal control inputs and parameters $\U^\ast$ minimizing the cost function of $\U$.
Even if an analytic expression of $\X(\U)$ does not exist, we can perform such optimization using a second-order method through \textit{sensitivity analysis} \cite{devincenti2021control, Zimmermann2019OptimalCV, bern2019trajectory}. SA allows us to compute the exact values of the first and the second derivatives efficiently. 

Firstly, We apply the chain rule to the cost function $\J(\X(\U),\,\U)$ for the total derivative:
\begin{equation}
\frac{\mathrm{d} \mathcal{J}}{\mathrm{d} \U} = \frac{\partial \mathcal{J}}{\partial \boldsymbol{X}} \frac{\mathrm{d} \boldsymbol{X}}{\mathrm{d} \U} + \frac{\partial \mathcal{J}}{\partial \boldsymbol{U}}\,.
\end{equation}
The partial derivative terms $\frac{\partial \mathcal{J}}{\partial \boldsymbol{X}}$ and $\frac{\partial \mathcal{J}}{\partial \boldsymbol{U}}$ are straightforward to compute. Meanwhile, the \emph{sensitivity} matrix $\Se \coloneqq \frac{\mathrm{d} \boldsymbol{X}}{\mathrm{d} \U} \in \mathbb{R}^{Nn \times (Nm+p)}$ requires additional steps for an analytic expression. 
%
According to the implicit function theorem, for a feasible pair $(\boldsymbol{X}, \U)$ that satisfies $\G(\X, \U) = \mathbf{0}_{Nn}$, \begin{equation}
    \frac{\mathrm{d} \G}{\mathrm{d} \U} = \frac{\partial \G}{\partial \X} \Se + \frac{\partial \G}{\partial \U} \eeq \mathbf{0}_{Nn \times (Nm+p)}\,.
\end{equation}
By rearranging the terms, we can express $\Se$ as: 
\begin{equation}\label{eq:sa}
    \Se = -\left(\frac{\partial \G}{\partial \boldsymbol{X}}\right)^{-1} \frac{\partial \G}{\partial \boldsymbol{U}}\,.
\end{equation}

Eventually, the analytic expression of the first and the second derivatives of the cost function are
\begin{subequations}
\label{eq:object-derivatives}
\begin{alignat}{1}
    \frac{\mathrm{d} \mathcal{J}}{\mathrm{d} \U} &= \frac{\partial \mathcal{J}}{\partial \X} \Se + \frac{\partial \mathcal{J}}{\partial \U}\,, \label{eqn:cost_derivative} \\
    \frac{\mathrm{d}^2 \mathcal{J}}{\mathrm{d} \U^2} &= \left(\frac{\mathrm{d}}{\mathrm{d} \boldsymbol{U}} \frac{\partial \mathcal{J}}{\partial \boldsymbol{X}} \right) \Se + \frac{\partial \mathcal{J}}{\partial \boldsymbol{X}} \frac{\mathrm{d} \Se}{\mathrm{d} \boldsymbol{U}} + \frac{\mathrm{d}}{ \boldsymbol{U}} \frac{\partial \mathcal{J}}{\partial \boldsymbol{U}} \label{eqn:cost_hessian} \\
    &\approx \Se^\top \frac{\partial^2 \mathcal{J}}{\partial \boldsymbol{X}^2} \Se + \Se^\top \frac{\partial^2 \mathcal{J}}{\partial \boldsymbol{U} \partial \boldsymbol{X}} + \frac{\partial^2 \mathcal{J}}{\partial \boldsymbol{X} \partial \boldsymbol{U}} \Se + \frac{\partial^2 \mathcal{J}}{\partial \boldsymbol{U}^2} \,. \label{eqn:gn_cost_hessian} 
\end{alignat}
\end{subequations}
For the full derivation of the second derivative, we refer the reader to the technical note by \citet{Zimmermann2019OptimalCV}. We note that the generalized Gauss-Newton approximation \eqref{eqn:gn_cost_hessian} can be employed in place of the Hessian \eqref{eqn:cost_hessian} to reduce the computational cost and to guarantee the semi-positive definiteness of the second derivative for nonlinear least-squares objectives.

\subsection{Quadrupedal Locomotion Control}
\label{sec:locomotion-control}

We formulate an OCP for quadrupedal locomotion using the framework described in \Cref{sec:framework}. The main objective is to track a reference base trajectory generated from a user's commands. Thus, we define the following cost function on the base positions $\r_k$ over a time horizon $N$:
\begin{subequations}
\label{eq:mpc-obj}
\begin{alignat}{1}
    \mathcal{J}(\X,\,\U) & \coloneqq K_1 \sum_{k=0}^{N} \|(\r_{k+1} - \r_k) - (\r_{k+1}^\refe - \r_k^\refe)\|_2^2 \label{eq:mpc-obj-body-traj} \\
    &+ K_2 \sum_{k=0}^{N} \|h_{k+1} - h_{k+1}^\refe\|_2^2 \label{eq:mpc-obj-body-height}\\ 
    &+ K_3 \sum_{i=1}^p \hspace{-0.2cm}\sum_{j=i+1}^{\min{(p, i+3)}} \|(\s^i - \s^j) - (\s^{\refe,i} - \s^{\refe,j})\|_2^2 \label{eq:mpc-obj-footstep-regularization} \\
    &+ \mathcal{R}_{\textrm{model}}(\X,\,\U) \,, \label{eq:mpc-obj-model-specifics}
\end{alignat}
\end{subequations}

The term \eqref{eq:mpc-obj-body-traj} penalizes base velocity tracking errors, \eqref{eq:mpc-obj-body-height} penalizes base height tracking errors, \eqref{eq:mpc-obj-footstep-regularization} regularizes the displacements between adjacent stepping locations, and finally \eqref{eq:mpc-obj-model-specifics} is a model specific cost term. The variable $\r_k$ is a part of the system state vector $\x_k$, and $\s^i$ is a part of the parameter vector $\p$. We provide the values of the weighting coefficients $K_i$ for all the cost terms we present in \Cref{tab:parameters}.

\Cref{eq:mpc-obj-footstep-regularization} regularizes the foothold optimization towards kinematically feasible solutions. We determine the reference footholds $\s^{\mathrm{ref},i}$ based on a simple \emph{impact-to-impact} method whereby support feet lie below the corresponding hip in the middle of the stance phase \cite{yin2021run}. We note that the term \eqref{eq:mpc-obj-footstep-regularization} only penalizes relative positions between stepping locations, thus making the corresponding support polygons loosely resemble the reference ones \cite{Xin2019OnlineRF}. 

\subsection{Examples}\label{subsec:examples}

We briefly describe two nonlinear systems, namely the variable-height inverted pendulum (IPM) and the single rigid body (SRBM) models, and we show how they can be integrated into our framework.

When possible, we discretize the continuous dynamics by employing a \emph{semi-implicit Euler method}; given $\r_k$ and $\r_{k-1}$, we approximate the velocity at time steps $k$ and $k+1$, respectively, as $\dot{\r}_k \approx (\r_k - \r_{k-1}) / \Delta t$ and $\dot{\r}_{k+1} \approx \dot{\r}_k + \ddot{\r}_k \Delta t$, where $\ddot{\r}_k$ can be computed using a model-specific dynamics equation:
\begin{align}
    \r_{k+1} \hspace{-0.06cm} &\approx \r_k + \dot{\r}_{k+1} \Delta t \nonumber \\
    &\approx \r_k + \dot{\mathbf{r}}_k \Delta t + \ddot{\mathbf{r}}_k \Delta t^2 \nonumber \\
    &\approx 2 \r_k - \r_{k-1} + \ddot{\r}_k \Delta t^2 \nonumber \\
    &= 2 \r_k - \r_{k-1} \nonumber \\ 
    & \qquad + \boldsymbol{f}_\textrm{model}(\r_k,\,\u_k,\,\s^{i_1},\,\s^{i_2},\,\ldots,\,\s^{i_{|\sigma_k|}}) \Delta t^2 \nonumber \\
    &\eqqcolon \mathbf{g}_{\textrm{model},k}\left(\r_{k-1},\,\r_k,\,\u_k,\,\s \right) \,, \label{eq:mpc-dynamics}
\end{align}
where $\sigma_k$ denotes the subset of the stance foot positions at time step $k$, and $\s^{i_j}\in\sigma_k,\,\forall j\in\{1,\,2,\,\ldots,\,|\sigma_k|\}$. The makeup of the control input vector $\u_k$ depends on the model and we will introduce it in due time.

In the following subsections, we define an explicit function $\boldsymbol{f}_\textrm{model}$ for the IPM and the SRBM. As mentioned in \Cref{sec:framework}, we define $\G_k \coloneqq \x_{k+1} - \mathbf{g}_k(\x_k,\,\u_k,\,\p)\,$ since an explicit expression of the system dynamics exists. 

\subsubsection{Inverted Pendulum Model}
\label{sec:inverted-pendulum}

\begin{figure} 
    \centering
    \includegraphics[width=\linewidth]{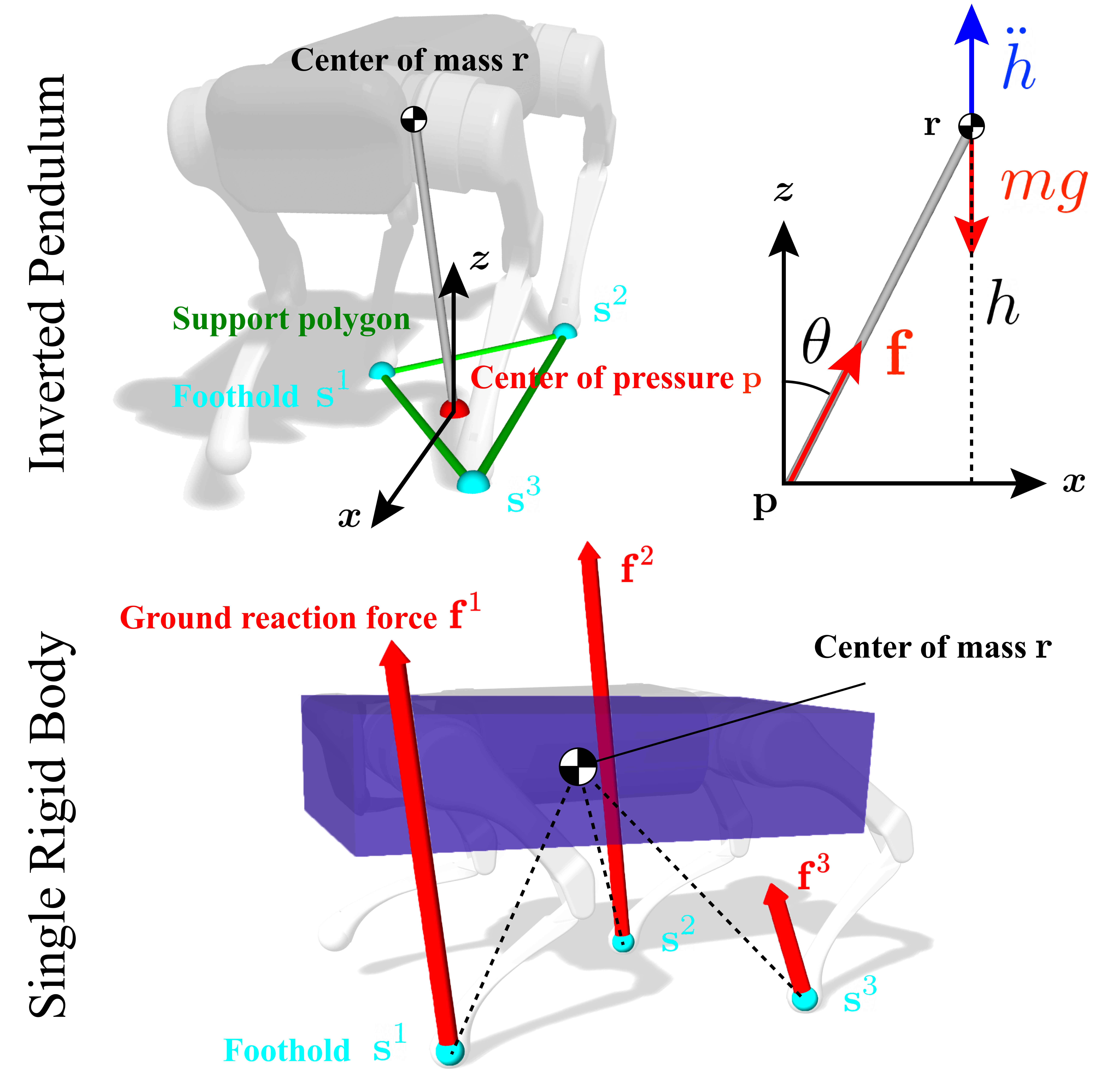}
    \caption{A quadrupedal robot represented as an inverted pendulum \textbf{(top)}, and a single rigid body \textbf{(bottom)}.} 
    \label{fig:nipm}
\end{figure}

\begin{table}
\caption{NMPC cost function parameter values.}
\label{tab:parameters}
\vspace{-0.1in}
\begin{center}
\setlength\tabcolsep{4.0pt}
\begin{tabular}{ccc|cc|cc|c|cc|c}
\hline
\multicolumn{3}{c|}{$\mathcal{J}$} & \multicolumn{2}{c|}{$\mathcal{R}_\textsc{IPM}$} & \multicolumn{2}{c|}{$\mathcal{R}_\textsc{SRBM}$} & $\mathcal{L}_\mathrm{gc}$ & \multicolumn{2}{c|}{$\mathcal{L}_\mathrm{ss}$} & $\mathcal{L}_\mathrm{ca}$ \\
\hline
$K_1$ & $K_2$ & $K_3$ & $K_4$ & $K_5$ & $K_6$ & $K_7$ & $K_8$ & $K_9$ & $K_{10}$ & $K_{11}$ \\
$1$ & $1$ & $0.2$ & $100$ & $1$ & $1$ & $1$ & $1$ & $0.1$ & $0.041$ & $1$ \\
\hline
\end{tabular}
\end{center}
\vspace{-0.1in}
\end{table}

The inverted pendulum model represents a legged robot as a point mass $m$ concentrated at the center of gravity of the system $\r$ and a massless telescoping rod in contact with a flat ground. We assume that the contact point of the rod is at the center of pressure (CoP) of the robot $\p$, i.e., the location at which the resultant ground reaction force vector $\mathbf{f}$ would act if it were considered to have a single point of application \cite{Cavanagh1978ATF}. The CoP always exists inside the support polygon of all stance foot positions $\s^i\in\R^3$. Thus, we can express its position with respect to an inertial reference frame as a convex combination:
\begin{equation}
\label{eqn:cop}
    \mathbf{p} = \sum_{\s^i\in\sigma} w^i \, \s^i,
\end{equation}
where $\sigma$ is the set of the stance foot positions, and $w^i\in\R_{\geq 0}$ is a non-negative scalar weight corresponding to $\s^i$ that satisfy $\sum_i w^i = 1 \,.$

The equation of motion for the IPM is given as follows: 
\begin{align}
    \ddot{\mathbf{r}} &= (\mathbf{r} - \sum_{\s^i\in\sigma} w^i \, \s^i) \frac{\ddot{h} + \|\mathbf{g}\|_2}{r_z} + \mathbf{g} \nonumber \\
    &\eqqcolon \boldsymbol{f}_\textsc{ipm}(\mathbf{r},\, \u,\,\s^{i_1},\, \s^{i_2},\, \ldots,\, \s^{i_{|\sigma|}}) \,, \label{eqn:ip_eom}
\end{align}
with control input vector $\u \coloneqq \left[ \ddot{h} \; w^{i_1} \; w^{i_2} \; \ldots \; w^{i_{|\sigma|}} \right]^\top$, and parameters $\s^{i_j}\in\sigma,\,\forall j\in\{1,\,2,\,\ldots,\,|\sigma|\}$. The derivation of \Cref{eqn:ip_eom} is available in the related paper \cite{kang2022animal}.

Furthermore, we define the model specific cost term \eqref{eq:mpc-obj-model-specifics} for the IPM as follows:
\begin{equation}
\label{eq:mpc-obj-ipmpc}
    \mathcal{R}_{\textsc{IPM}} \eqqcolon \sum_{k=0}^{N-1} \left( \frac{K_4}{2} \| 1 - \sum_i w_k^i \|_2^2 + K_5 \sum_i \mathcal{S}_{\geq 0}( w_k^i ) \right) \hspace{-0.1cm}\,, 
\end{equation}
where $\mathcal{S}_{\geq r}\colon\R\rightarrow\R_{\geq 0},\,\forall r\in\R$ is a $\mathcal{C}^2$-continuous function following [\citenum{Bern2017InteractiveDO}, eq. (8)], namely
\begin{equation*}
    \mathcal{S}_{\geq r}(x) \coloneqq
\begin{cases}
0 &\Gamma \geq \epsilon \\ 
-\frac{1}{6\epsilon}\Gamma^3 + \frac{1}{2}\Gamma^2 + \frac{\epsilon}{2}\Gamma + \frac{\epsilon^2}{6} &-\epsilon \leq \Gamma < \epsilon \\ 
\Gamma^2 + \frac{\epsilon^2}{3} &\Gamma < -\epsilon 
\end{cases}
\end{equation*}
with $\Gamma = x-r$ and $\epsilon = 0.1$. This term enforces the constraint $\sum_i w^i = 1 \,$, and the non-negativity of the weights as a soft constraint. 


\subsubsection{Single Rigid Body Model}

If the limbs of a robot are lightweight compared to its body, we can neglect their inertial effects and reduce the system to a single rigid body with mass $m$ and body frame moment of inertia ${}^B\mathbf{I}\in\mathbb{R}^{3\times 3}$.  
The position $\r\in\mathbb{R}^3$ and unit quaternion $\mathbf{q}\in\mathbb{S}^3$ define the pose of the lumped rigid body. ${}^B\boldsymbol{\omega}$ denotes its angular velocity vector expressed in body frame, and $\mathbf{f}^i\in\mathbb{R}^3$ denotes the ground reaction force associated with the stepping location $\mathbf{s}^i\in\sigma$. Then, we can write the dynamics of the system as:
\begin{align}
    \ddot{\r} &= \frac{1}{m} \sum_{\mathbf{s}^i\in\sigma} \mathbf{f}^i + \mathbf{g} \eqqcolon \boldsymbol{f}_{\textsc{srbm}, \mathrm{t}}(\u) \,, \label{eq:srbm-lin-dynamics} \\
    {}^B\dot{\boldsymbol{\omega}} &= {}^B\mathbf{I}^{-1} \left[ \mathbf{R}(\mathbf{q})^\top \sum_{\mathbf{s}^i\in\sigma} \left( \mathbf{s}^i - \r \right)\cross\mathbf{f}^i - {}^B\boldsymbol{\omega}\cross{}^B\mathbf{I}{}^B\boldsymbol{\omega} \right] \nonumber \\
    &\eqqcolon \boldsymbol{f}_{\textsc{srbm}, \mathrm{r}}(\r,\, \q,\, {}^B\boldsymbol{\omega},\, \u,\, \s^{i_1},\, \s^{i_2},\, \ldots,\, \s^{i_{|\sigma|}}) \,, \label{eq:srbm-rot-dynamics}
\end{align}
where $\mathbf{R}(\mathbf{q})$ is the rotation matrix corresponding to $\mathbf{q}$, and $\u \coloneqq \left[ \mathbf{f}^{i_1} \; \mathbf{f}^{i_2} \; \ldots \; \mathbf{f}^{i_{|\sigma|}} \right]^\top$ is the control input vector.

We employ a semi-implicit Euler method similar to the one outlined in \Cref{subsec:examples}. However, to integrate the orientation dynamics, we employ a forward Lie-group Euler method which allows us to preserve the unitary norm constraint of unit quaternions. Specifically, we approximate the body frame angular velocity at time step $k$ and $k+1$, respectively, as ${}^B\boldsymbol{\omega}_k \approx 2\,\mathfrak{Im}(\bar{\mathbf{q}}_{k-1} \ast \mathbf{q}_k) / \Delta t$ and ${}^B\boldsymbol{\omega}_{k+1} \approx {}^B\boldsymbol{\omega}_k + {}^B\dot{\boldsymbol{\omega}}_k \Delta t$, where $\mathfrak{Im}(\q)$ extracts the imaginary part of $\q$, $\bar{\q}$ is the conjugate of $\q$, $\ast$ is the quaternion multiplication operator, and ${}^B\dot{\boldsymbol{\omega}}_k$ can be computed using \eqref{eq:srbm-rot-dynamics}. Then, our integration scheme for unit quaternions translates to $\mathbf{q}_{k+1} \approx \q_k \ast \exp\left({}^B\boldsymbol{\omega}_{k+1} \Delta t\right)$, where $\exp\colon\mathbb{R}^3\rightarrow\mathbb{S}^3$ is a Lie-group exponential function which, for unit quaternions, has the following closed form:
\begin{equation*}
    \exp(\mathbf{v}) \coloneqq 
\begin{cases}
\cos(\frac{1}{2}\|\mathbf{v}\|) + \frac{\mathbf{v}}{\|\mathbf{v}\|} \sin(\frac{1}{2}\|\mathbf{v}\|) &\|\mathbf{v}\| \neq 0 \\ 
1  &\|\mathbf{v}\| = 0
\end{cases} \,.
\end{equation*}

Using the equations above, we can finally write the SRBM dynamics in the form \eqref{eq:mpc-dynamics} as:
\begin{align}
    \begin{bmatrix}
    \r_{k+1} \\ 
    \q_{k+1}
    \end{bmatrix} &\approx \begin{bmatrix}
    2 \r_k - \r_{k-1} + \boldsymbol{f}_{\textsc{srbm}, \mathrm{t}}(\u_k) \Delta t^2 \\
    \q_k \ast \exp({}^B\boldsymbol{\omega}_{k+1} \Delta t )
    \end{bmatrix} \nonumber \\
    &\coloneqq \mathbf{g}_{\textsc{srbm},k}(\r_{k-1},\, \r_k,\, \q_{k-1},\, \q_k,\, \u_k,\, \s) \,. \label{eq:srbm-dynamics}
\end{align}

Following \eqref{eq:mpc-obj-model-specifics}, we can define a cost term for the SRBM penalizing deviations from a reference orientation trajectory $\q_k^\refe$ \cite{Jackson2021PlanningWA} and imposing non-negative vertical components of the GRFs\footnote{For our preliminary results, we do not include friction cone soft constraints to \eqref{eq:mpc-obj-srbm} to keep our implementation simple.}:
\begin{equation}
\label{eq:mpc-obj-srbm}
    \mathcal{R}_{\textsc{SRBM}} \eqqcolon \sum_{k=0}^{N-1} \left( K_6 \left( 1-\left| \q_k^\top \q^{\refe}_k \right| \right) + K_7 \sum_i \mathcal{S}_{\geq 0}( \mathbf{f}_{z, k}^i ) \right) \hspace{-0.1cm}\,.
\end{equation}

\section{Experiments}

We present a series of simulation and hardware experiments we conducted to verify the efficacy of our approach. The results we discuss in this section were attained using the IPM described in \Cref{sec:inverted-pendulum}. The footage of the experiments is available in the supplementary video\footnote[2]{The video is available in \url{https://youtu.be/BrJSRlAJaX4}.}, along with preliminary results achieved with the SRBM. In all our experiments, the optimal base trajectories output by the MPC scheme were tracked by a quadratic programming-based whole-body controller \cite{devincenti2021control}.

\begin{figure} 
    \centering
    \includegraphics[width=\linewidth]{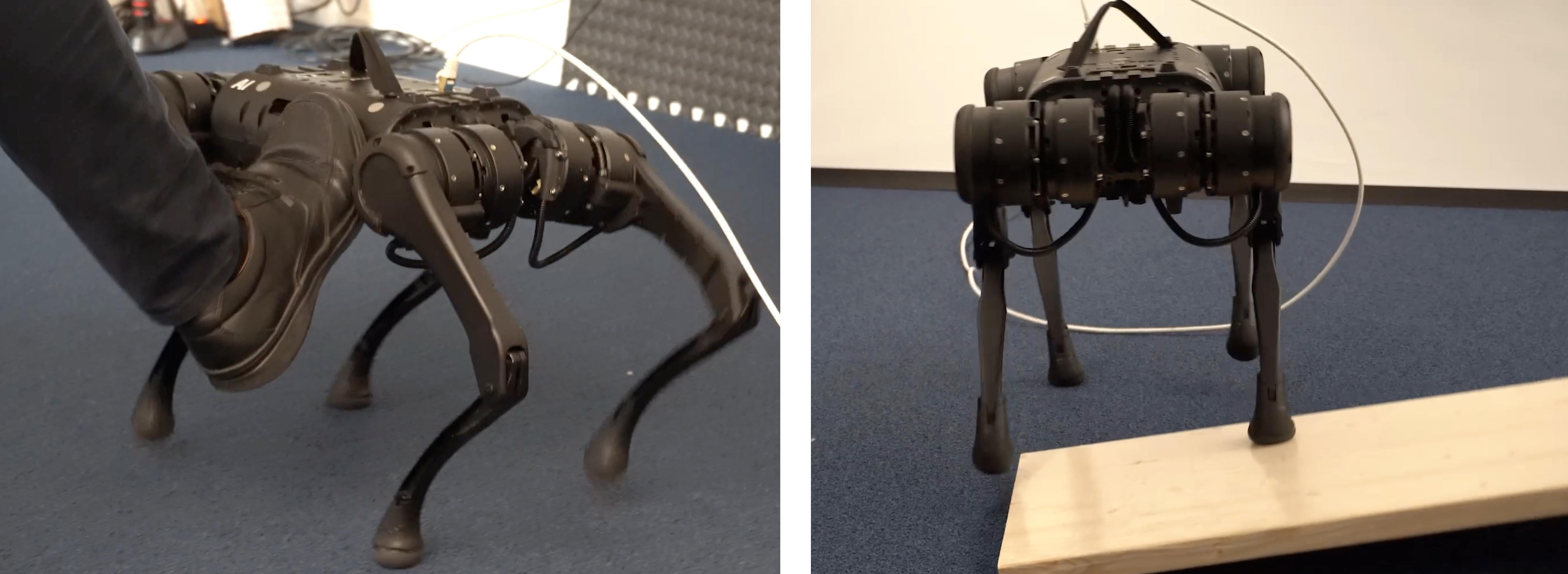}
    \caption{We pushed \textbf{(left)} and disturbed the \emph{Unitree A1} robot by putting a plate under its feet \textbf{(right)} while it was performing a trot gait.} 
    \label{fig:disturbance}
\end{figure}

Firstly, we tested the robustness of our controller for locomotion on flat terrains using the \textit{Unitree A1} robot. We hindered the robot while it was trotting in place, as portrayed in the snapshots in \Cref{fig:disturbance}. In our tests, the robot was able to withstand unexpected disturbances and successfully recover its stability. We demonstrate in the accompanying video how the foothold optimization improves the capability of the system to resist large lateral pushes.

\begin{figure} 
    \centering
    \includegraphics[width=\linewidth]{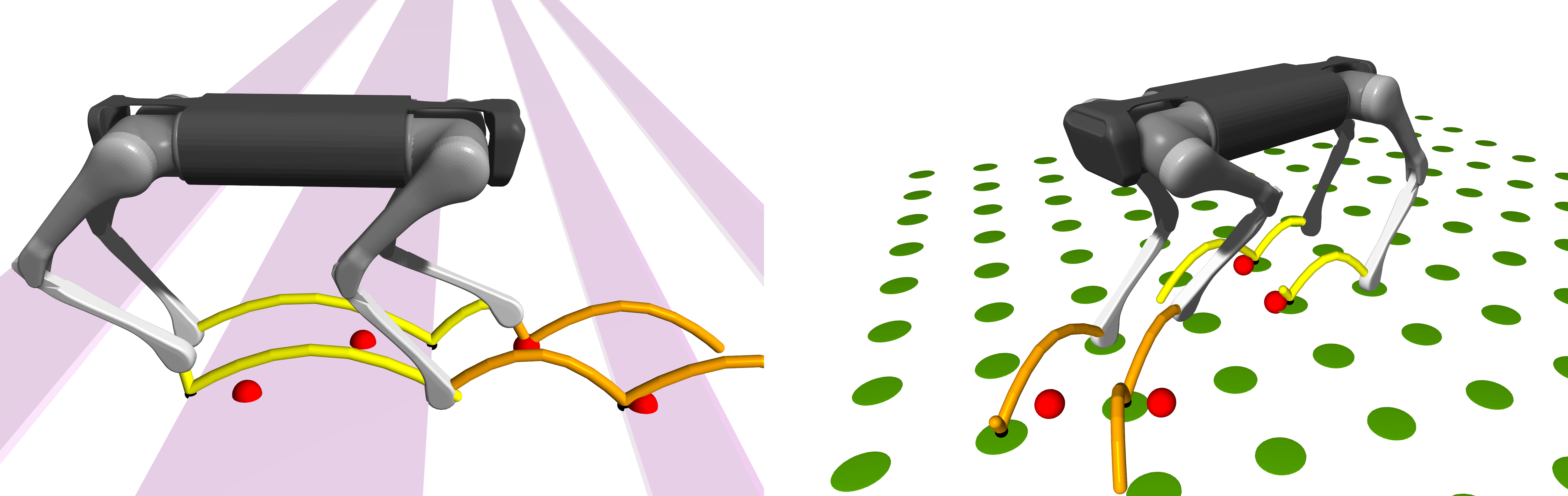}
    \caption{Snapshots of simulation experiments for the gap crossing (\textbf{left}) and stepping stone (\textbf{right}) scenarios with the \textit{Aliengo} robot. The \textbf{red} spheres depict the reference stepping locations, while the \textbf{yellow} and \textbf{orange} trajectories are the outputs of our MPC controller. The resulting footstep placements deviate considerably from the corresponding references and allow the robot to avoid \SI{32}{\centi\meter} wide gaps (\textbf{left, in light red}) and step on isolated footholds (\textbf{right, in green}).} 
    \label{fig:gc-ss}
\end{figure}

We show the versatility of our foothold optimization approach to adapt a gait to different terrain types, namely a \emph{gap crossing} and a \emph{stepping stones} scenarios. The former setting consists of a sequence of rifts with different widths; the latter comprises a grid of stepping stones distant \SI{20}{\centi\meter} from each other the robot must step on -- see \Cref{fig:gc-ss}.
We add the following terms to the objective function \eqref{eq:mpc-obj} to model each gap and stepping stone, respectively:
\begin{align}
    \mathcal{L}_\mathrm{gc}(\s) &\coloneqq K_8 \sum_{i=1}^p \mathcal{S}_{\geq g}(| s_x^i - g_x |) \,, \label{eq:gap-penalty} \\
    \mathcal{L}_\mathrm{ss}(\s) &\coloneqq K_9 \sum_{i=1}^p -\exp{-\frac{1}{2} \frac{\|\s^i - \mathbf{t}\|_2^2}{K_{10}^2}} \,, \label{eq:ss-penalty}
\end{align}
where $g$ and $g_x$ are the gap half width and \textit{x}-position, respectively, $\mathbf{t}$ is the stepping stone location, and $K_9$ and $K_{10}$ are tuning parameters. To model the stepping stones in \eqref{eq:ss-penalty}, we employ a negative Gaussian function centered at the corresponding positions; in this way, we incentivize nearby stepping locations to converge towards the closest footholds.
As shown in the supplementary video, these simple penalty terms are sufficient to ensure that the associated constraints are almost never violated. The occasional missteps may be avoided through a careful tuning of \eqref{eq:gap-penalty} and \eqref{eq:ss-penalty}, or by designing some fallback control strategies. 

\begin{figure} 
    \centering
    \includegraphics[width=\linewidth, clip, trim=0cm 3.6cm 0cm 0cm]{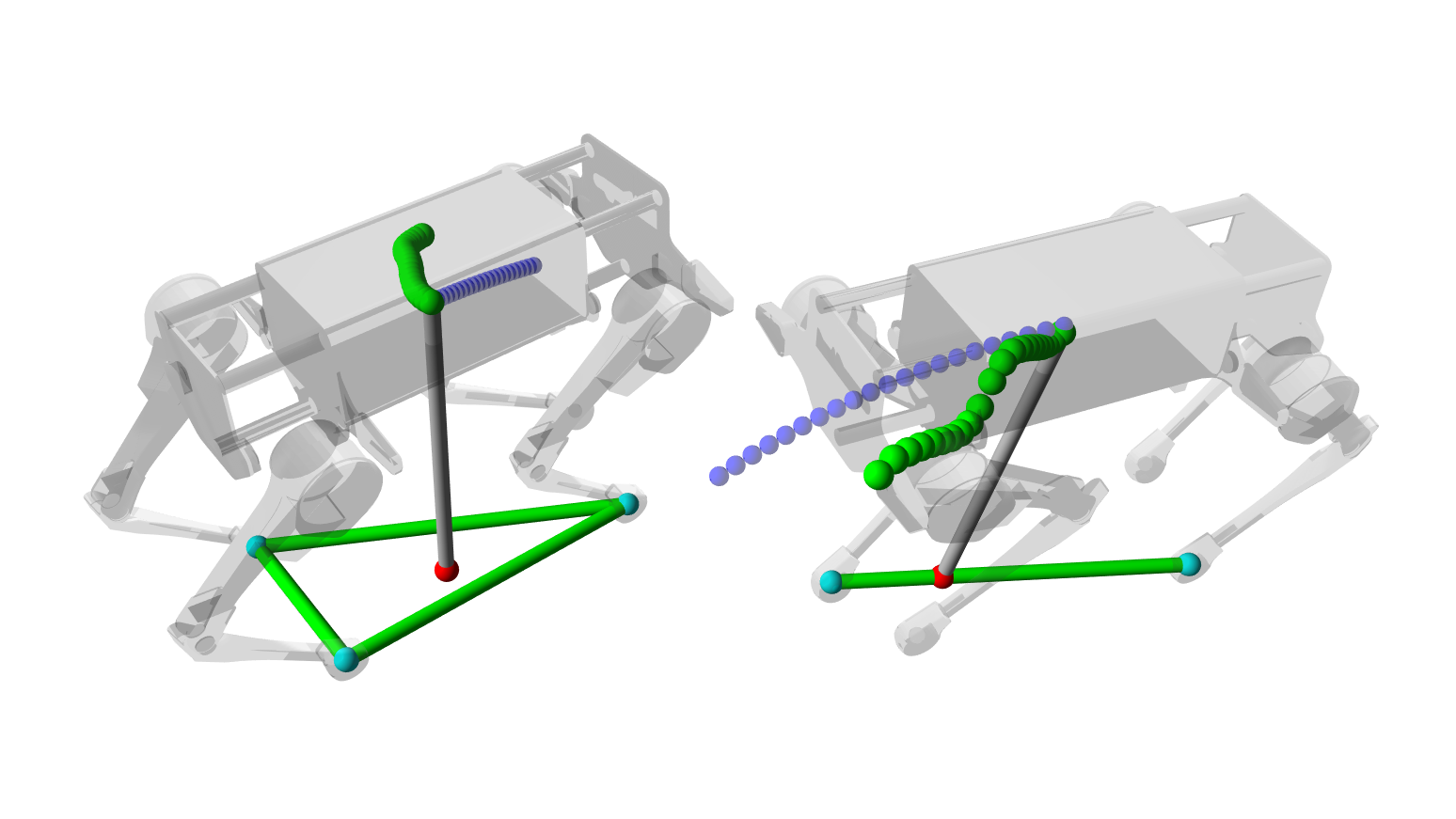}
    \caption{Two \textit{Laikago} quadruped robots controlled using a centralized MPC strategy in simulation. The robot on the left is executing a walking gait, whereas the one on the right is performing a flying trot. The reference trajectories are denoted by blue spheres, and the optimized ones are represented by green spheres: the latter deviate significantly from the former to prevent the robots from colliding.} 
    \label{fig:ipm-pair}
\end{figure}

Finally, we extend our framework so that the state vector contains the positions of two robots, and we couple the solutions for the two subsystems by adding the following collision avoidance term to the objective function:
\begin{equation*}
    \mathcal{L}_\mathrm{ca}(\X) \coloneqq K_{11} \sum_{k=0}^N\mathcal{S}_{\geq 1}(\|\r_k^a - \r_k^b\|_2) \,,
\end{equation*}
where $\r_k^a$ and $\r_k^b$ are the states of the two robots at time step $k$, respectively. This cost term ensures that the robots keep a distance of at least \SI{1}{\meter} from each other -- see \Cref{fig:ipm-pair}. As shown in the supplementary video, our NMPC framework is able to control the multi-robot system in real time by solving a single OCP.


\section{Conclusion and Future Work}

Our NMPC scheme facilitates the implementation of robust controllers for various quadrupedal locomotion tasks. We can easily integrate different nonlinear dynamics models into our framework. We leave a complete demonstration with different models and more comprehensive analysis for future work. Our immediate next step is to verify our formulation of the SRBM and test it on hardware. Furthermore, we intend to compare our method to other state-of-the-art nonlinear control frameworks.

\addtolength{\textheight}{-12cm}   



%





\bibliographystyle{IEEEtranN}
\bibliography{root.bib}

\end{document}